\documentclass[sigconf]{acmart}
\usepackage{subfigure}
\usepackage{graphicx}
\usepackage{verbatim}
\usepackage{arydshln}

\usepackage{color} 
\AtBeginDocument{%
  }

\copyrightyear{2022}
\acmYear{2022}
\setcopyright{acmcopyright}\acmConference[MM '22]{Proceedings of the 30th ACM International Conference on Multimedia}{October 10--14, 2022}{Lisboa, Portugal}
\acmBooktitle{Proceedings of the 30th ACM International Conference on Multimedia (MM '22), October 10--14, 2022, Lisboa, Portugal}
\acmPrice{15.00}
\acmDOI{10.1145/3503161.3548093}
\acmISBN{978-1-4503-9203-7/22/10}

\begin{document}

\title{Skeleton2Humanoid: Animating Simulated Characters for Physically-plausible Motion In-betweening}

\author{Yunhao Li}
\authornotemark[1]
\affiliation{%
  \institution{Institute of Image Communication and Network Engineering}
  \institution{Shanghai Jiao Tong University}
  \city{Shanghai}
  \country{China}
}
\email{lyhsjtu@sjtu.edu.cn}

\author{Zhenbo Yu}
\authornote{Equal contribution.}
\affiliation{%
  \institution{Shanghai Jiao Tong University}
  \city{Shanghai}
  \country{China}
  }
\email{yuzhenbo@sjtu.edu.cn}

\author{Yucheng Zhu}
\affiliation{%
  \institution{Institute of Image Communication and Network Engineering}
  \institution{Shanghai Jiao Tong University}
  \city{Shanghai}
  \country{China}
}
\email{zyc420@sjtu.edu.cn}

\author{Bingbing Ni}
\affiliation{%
  \institution{Shanghai Jiao Tong University}
  \city{Shanghai}
  \country{China}
}
\email{nibingbing@sjtu.edu.cn}

\author{Guangtao Zhai}
\authornotemark[2]
\affiliation{%
  \institution{Institute of Image Communication and Network Engineering}
  \institution{Shanghai Jiao Tong University}
  \city{Shanghai}
  \country{China}
}
\email{zhaiguangtao@sjtu.edu.cn}

\author{Wei Shen}
\authornote{Corresponding author.}
\affiliation{%
  \institution{MoE Key Lab of Artificial Intelligence, AI Institute}
  \institution{Shanghai Jiao Tong University}
  \city{Shanghai}
  \country{China}
}
\email{wei.shen@sjtu.edu.cn}

\renewcommand{\shortauthors}{Yunhao Li et al.}

\begin{abstract}
 Human motion synthesis is a long-standing problem with various applications in digital twins and the Metaverse. However, modern deep learning based motion synthesis approaches barely consider the physical plausibility of synthesized motions and consequently they usually produce unrealistic human motions. In order to solve this problem, we propose a system ``Skeleton2Humanoid'' which performs physics-oriented motion correction at test time by regularizing synthesized skeleton motions in a physics simulator. Concretely, our system consists of three sequential stages: (I) test time motion synthesis network adaptation, (II) skeleton to humanoid matching and (III) motion imitation based on reinforcement learning (RL). Stage I introduces a test time adaptation strategy, which improves the physical plausibility of synthesized human skeleton motions by optimizing skeleton joint locations. Stage II performs an analytical inverse kinematics strategy, which converts the optimized human skeleton motions to humanoid robot motions in a physics simulator, then the converted humanoid robot motions can be served as reference motions for the RL policy to imitate. Stage III introduces a curriculum residual force control policy, which drives the humanoid robot to mimic complex converted reference motions in accordance with the physical law. We verify our system on a typical human motion synthesis task, motion-in-betweening. Experiments on the challenging LaFAN1 dataset show our system can outperform prior methods significantly in terms of both physical plausibility and accuracy. Code will be released for research purposes at: \url{https://github.com/michaelliyunhao/Skeleton2Humanoid}.

\end{abstract}


\begin{teaserfigure}
  \centering
  \includegraphics[scale = 0.90]{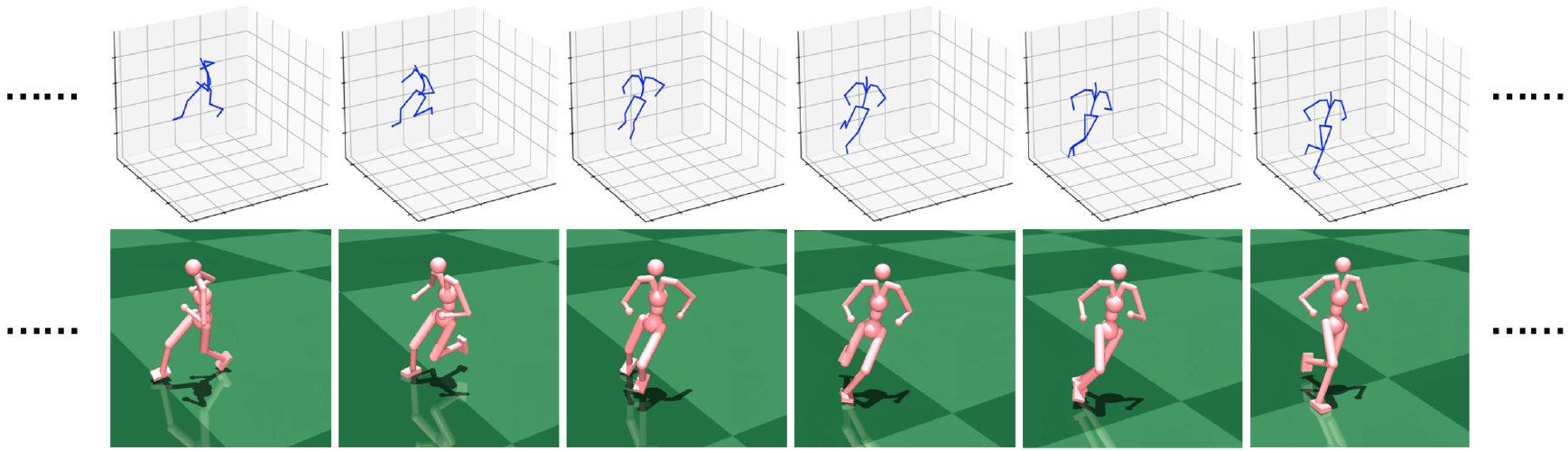}
  \caption{Our Skeleton2Humanoid system can directly synthesize a complete humanoid character transition motion in a physics simulator (\bf{Bottom}) given past keyframes and a future keyframe (\bf{Top}). Our system can produce both accurate and physically-plausible character motions.}
  \label{fig:teaser}
\end{teaserfigure}

\maketitle

\section{Introduction}
Synthesizing both accurate and realistic virtual human motions has been a widely explored but challenging task in computer vision and graphics \cite{min2009interactive,kovar2008motion} with various applications in digital twins and the Metaverse. Recently, deep learning sheds light onto a way to generate accurate human motions and has been applied to various motion synthesis tasks, such as human motion prediction \cite{butepage2017deep,fragkiadaki2015recurrent,martinez2017human,li2017auto,zhang2021we,yuan2020dlow,mao2020history,mao2019learning,jain2016structural,aksan2019structured}, human motion completion \cite{kaufmann2020convolutional,wang2021synthesizing,hernandez2019human} and human motion in-betweening \cite{harvey2020robust,duan2022unified,cai2021unified,zhou2020generative}. Although they have shown great performance on synthesizing accurate human body motions with small skeleton joint errors comparing with ground truth motions, they fail to model the motions under the physics laws. Consequently, the synthesized motions are usually physically implausible. For example, the synthesized feet often penetrate the ground, the body joints are rotated with impossible angles, the whole body motions are unsmooth, the synthesized feet slide back and forth while they should be in static and touch the ground. These synthesized artifacts significantly limits the application of motion synthesis on the virtual human animation and the incoming Metaverse because they easily make humans feel unrealistic.

Utilizing humanoid characters in a physics simulator to optimize motions is a promising solution because the physics simulator can guarantee the physical plausibility of the generated motions. Prior works \cite{peng2018deepmimic,peng2018sfv,wang2020unicon} utilized reinforcement learning (RL) to actuate the humanoid character to imitate various reference mocap data for creating physical character animation. Inspired by them, Recent works \cite{yuan2021simpoe,luo2021dynamics} also attempted to utilize RL to imitate motions synthesized by deep neural networks, in the format of skeletons or SMPL \cite{loper2015smpl} models, aiming at producing physically-plausible motions for 3D pose estimation. However, these methods are only validated on simple motions such as walking and talking in the Human3.6m dataset and cannot generalize well to complex motions or irregular motions. In addition, RL based imitation requires transferring synthesized human skeleton motions to humanoid motions, where a humanoid character should be carefully designed to exactly match the human skeletons in terms of both shapes and the kinematics tree. This limits RL based imitation to transfer motions between skeleton and humanoid with different shapes and kinematics trees.

To address these issues, we propose Skeleton2Humanoid, a novel system which is able to improve the physical plausibility of the motions synthesized from motion synthesis networks, though the transfer from human skeleton motions to humanoid character motions. Our Skeleton2Humanoid system consists of three sequential stages: \textbf{(I)} \textbf{Test Time Motion Synthesis Network Adaptation:} We adapt the motion synthesis network with a few gradients on the test data using two new self-supervised losses, a foot contact consistency loss and a motion smoothness loss, which can improve the physical plausibility of the predicted motions. \textbf{(II)} \textbf{Skeleton to Humanoid Matching:} We match the synthesized human skeleton motions to humanoid character motions by a novel general analytical inverse kinematic method. Inverse kinematics is able to convert human skeleton motions to humanoid motions even when the body structure is different from the human skeleton. \textbf{(III)} \textbf{Motion Imitation base on RL:} Finally, we animate the humanoid character to imitate various synthesized motions. Specifically, based on recent work \cite{luo2021dynamics, yuan2020residual}, we propose a curriculum residual force control humanoid control policy (CRP) by introducing a curriculum learning paradigm that dynamically adjusts a residual force scale during RL training, which can improves asymptotic RL performance on imitating various synthesized motions. 
To verify the effectiveness of our Skeleton2Humanoid system, we select ``motion in-betweening'' task, as it is a recent proposed challenging motion prediction task \cite{harvey2020robust, duan2022unified} for evaluation. Motion in-betweening aims at predicting the transition motions between the past given keyframes and a provided future keyframe. Experiments
on challenging LaFAN1 dataset show the superiority of our Skeleton2Humanoid system.

The main contributions of this paper are as follows:
\textbf{(1)} We present Skeleton2Humanoid, a new system that converts human skeleton motions to humanoid character motions to produce physical plausible motions. \textbf{(2)} Our proposed test time adaptation stage can further improve the prediction accuracy and physical plausibility on large mocap dataset LaFAN1 for the motion in-betweening task. With test time adaptation, we achieve a new benchmark accuracy on the motion in-betweening task. \textbf{(3)} Our proposed curriculum residual force control policy enables finer character control and outperforms prior arts on motion imitation. \textbf{(4)} Our whole Skeleton2Humanoid system significantly improves the performance of human in-betweening motions on physical plausibility and achieves comparable motion prediction accuracy.

\begin{figure*}[h]
\centering
\includegraphics[scale=0.92]{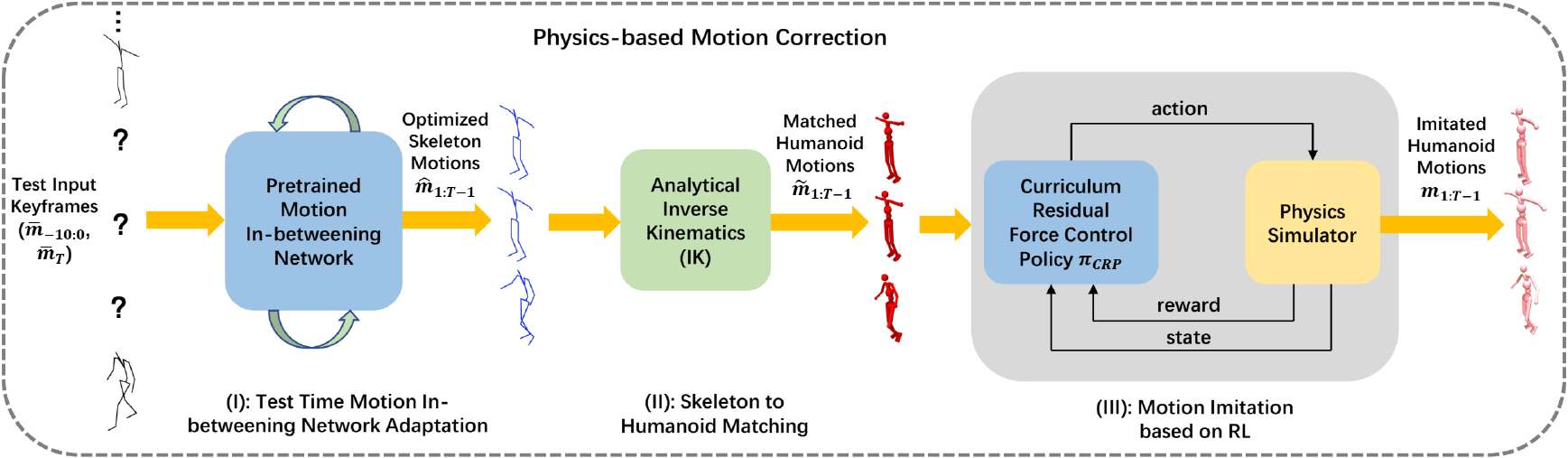}
\caption{An overview of our Skeleton2Humanoid system on the motion in-betweening task. Given the test data containing past keyframes $\overline{m}_{-9:0}$ and a future keyframe $\overline{m}_{T}$, Stage I optimizes skeleton joint locations by test time adaptation and produces more plausible skeleton motions $\hat{m}_{1:T-1}$. Stage II converts the optimized skeleton motions $\hat{m}_{1:T-1}$ to humanoid motions $\tilde{m}_{1:T-1}$ in the physics simulator by analytical inverse kinematics. Stage III finally drives the humanoid to mimic converted skeleton motions $\tilde{m}_{1:T-1}$ to produce physically-plausible humanoid motions ${m}_{1:T-1}$.}
\label{skeleton2humanoid}
\vspace{-2mm}
\end{figure*}
\vspace{-3mm}


\section{Related Work}
\noindent\textbf{Human/character motion synthesis}: Motion synthesis is a general term which contains several tasks including motion prediction, in-betweening and completion. Motion prediction aims at predicting future human motions given past motions. Deterministic motion prediction estimates a single accurate motion and prior works used various network architectures including recurrent neural network \cite{fragkiadaki2015recurrent,martinez2017human,jain2016structural}, graph convolution network \cite{li2020dynamic} or transformer \cite{mao2020history} to model human motions. Stochastic motion prediction produces diverse future human motions by utilizing generative model such as VAE \cite{yuan2020dlow,zhang2021we,guo2020action2motion,petrovich2021action}, GAN \cite{barsoum2018hp,gui2018adversarial,song2022actformer}. Motion completion and in-betweening aim at filling gaps of motion with predefined key-frame constraints. Current works utilized convolution networks \cite{yan2019convolutional,kaufmann2020convolutional,hernandez2019human,zhou2020generative}, recurrent networks \cite{harvey2020robust,cai2018deep} or transformers \cite{duan2022unified} to synthesize accurate and consistent results. For instance, Harvey et al. \cite{harvey2020robust} proposed a transition generation technique based on recurrent neural networks for motion in-betweening task. Duan et al. \cite{duan2022unified} utilized transformer architecture to model human motions in a sequence-to-sequence manner for the motion in-betweening task.

\noindent\textbf{Test Time Adaptation}: Test time adaptation is a recently proposed method that utilize the self-supervised distribution information from the test data presented at test
time to quickly adapt models with a few gradient steps \cite{sun2020test, pandey2021generalization, wang2020tent}, which can further improve the model performance on test data. The first work \cite{sun2020test} introduced test time adaptation by proposing an auxiliary branch with self-supervised rotation prediction loss to adapt the classification model. Wang et al. \cite{wang2020tent} minimized the predicted entropy of classification model on test data to improve the performance. Recently, more works start to utilize test time adaptation on the 2D/3D human pose related task \cite{li2021test, guan2021bilevel}: For instance, Guan et al. \cite{guan2021bilevel} proposed an online bilevel adaptation framework for 3D human mesh reconstruction which greatly improves model generalization. In contrast to other works, Our approach is the first one to study test time adaptation on the human motion in-betweening task.

\noindent\textbf{Reinforcement Learning for Humanoid Character Control}: Deep RL is a promising approach for learning character control policies \cite{liu2017learning, liu2018learning, peng2018deepmimic, peng2018sfv, ho2016generative} to help character perform various motions. Peng et al. \cite{peng2018deepmimic} first utilize hand craft rewards to imitate a single sequence of human poses. Recently, some works \cite{gong2022posetriplet,yuan2021simpoe,yuan20183d,luo2021dynamics,yuan2019ego} used RL to produce simple human motions from egocentric videos for ego-pose estimation or 3d human pose estimation. Yuan et al. \cite{yuan2020residual} proposed to add external residual forces and help characters to better imitate agile single reference motions. In addition, some works \cite{bergamin2019drecon, park2019learning, won2020scalable} utilized deep RL to learn a interactive controllable policies from large motion capture data for character animation. However, Prior works mostly focused on learning control policies on motion capture data, while we learn a policy to imitate synthesized motions. We propose a curriculum residual force control policy (CRP) that can better imitate diverse motions.


\section{Approach}
\subsection{System Overview}
The human motion in-betweening task can be formulated as: given the past 10 human skeleton poses $\overline{m}_{-9:0}$ and a future skeleton keyframe $\overline{m}_{T}$ at time T, we want to recover the ground truth transition motions $\overline{m}_{1:T-1}$. Given a pretrained typical motion in-betweening network \cite{harvey2020robust}, our Skeleton2Humanoid performs a physics-oriented motion correction consists of 3 stages as presented in Fig. \ref{skeleton2humanoid} to optimize synthesized in-betweening motions. Stage I optimizes the pretrained motion in-betweening network at test time to predict more physically-plausible skeleton transition motions $\hat{m}_{1:T-1}$. Then Stage II transfers the optimized skeleton motions $\hat{m}_{1:T-1}$ to humanoid motions $\tilde{m}_{1:T-1}$ through analytical inverse kinematics. Finally, Stage III learns a curriculum residual force control policy to imitate the transferred humanoid motions $\tilde{m}_{1:T-1}$ to produce physically-plausible humanoid motions ${m}_{1:T-1}$.

In our Skeleton2Humanoid framework, $\overline{m}_t$ and $\hat{m}_t$ are skeleton motions, and $\overline{m}_t$ is represented by $\overline{m}_t \triangleq (\overline{q}_t, \overline{r}_t, \overline{p}_t)$, where $\overline{q}_t$ and $\overline{r}_t$ denote body joint angles in quaternions and root translation, $\overline{p}_t$ denotes 3d joint positions calculated by forward kinematics. Similarly, $\hat{m}_t \triangleq (\hat{q}_t, \hat{r}_t, \hat{p}_t)$. In addition, $\tilde{m}_{t}$ and ${m}_{t}$ are humanoid motions, $\tilde{m}_{t}$ is represented by $\tilde{m}_{t} \triangleq (\tilde{q}_{t}, \tilde{r}_{t}, \tilde{p}_{t})$, where $\tilde{q}_{t}$, $\tilde{r}_{t}$ and $\tilde{p}_{t}$ denote joint angles in euler angles, root translation and 3d joint positions of the reference humanoid motions. Similarly, ${m}_{t} \triangleq ({q}_{t}, {r}_{t}, {p}_{t})$.

\begin{figure}[t]
\centering
\includegraphics[scale=0.50]{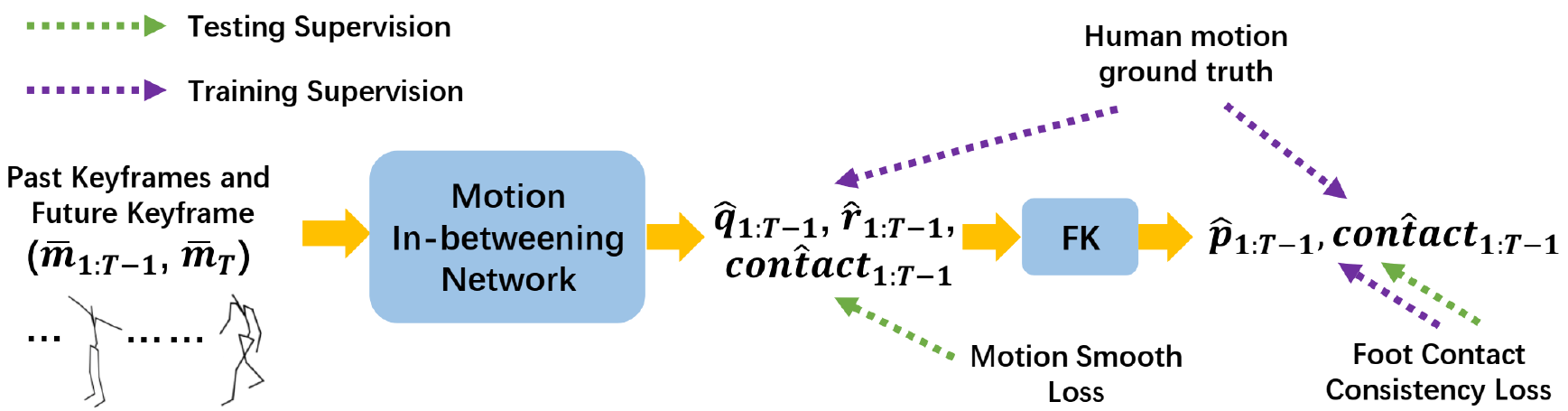}
\caption{Details of our test time adaptation method. $\hat{q}$, $\hat{contact}$, FK and $\hat{p}$ represent the predicted root positions, the contact prediction of feet joints, the forward kinematic process and 3d joint positions for human skeleton, respectively.}
\label{fig:ttt}       
\vspace{-2mm}
\end{figure}
\vspace{-2mm}

\subsection{Test Time Motion In-betweening Network Adaptation}
\subsubsection{Adaptation for Physically-plausible Skeleton Motion}
Previous human motion in-betweening model \cite{harvey2020robust} has shown great performance on synthesizing accurate human motions. However, it still suffers from implausibility on testing data because of the domain shift between training data and testing data. For example, the synthesized human motions usually have the foot sliding problem which makes motions look strange and it is hard for humanoid in a physics simulator to imitate this motion. In addition, the root joint velocity of the synthesized human motion sometimes changes drastically and results in on unsmooth motion. Inspired by these observations, we propose a test time adaptation method which updates the motion in-betweening network on test data and design two new self-supervised losses for test time adaptation. 

\noindent\textbf{Foot contact consistency loss: } We encourage that the feet should be in static contact with the ground when performing various motions. The loss is:
\begin{equation}
L_{contact} = \sum_{t=0}^{t=T-2}\sum_{foot\in F}^{}  \left |  \hat{p}^{foot}_{t+1} - \hat{p}^{foot}_{t}\right | \cdot \hat{contact}_{t+1} ^{foot},
\end{equation}
where $\hat{P}^{foot}_t$ is the predicted 3D position of the foot joint at time t, $F$ is a subset of joints which contains all foot-related joints, i.e. "left ankle", "right ankle", "left front feet" and "right front feet". $\hat{contact}_{t+1} ^{foot}$ represents the predicted contact probability for the foot joint at time t.

\noindent\textbf{Motion smoothness loss:} We encourage the root joint positions along the time axis to be smooth. The motion smoothness loss is:
\begin{equation}
L_{smooth} =  \sum_{t=0}^{T-2} \left \|\hat{r}_{t+1} - \hat{r}_{t} \right \| _2,
\end{equation}
where $r_{t}$ is the 3D position of root joint at time t.

By utilizing these two losses, we can optimize the motion in-betweening network which is trained on the training set given the test set. In the following section, we will introduce our method in detail.

\subsubsection{Implementation Detail}
Now, we describe the training phase and the testing phase of our test time adaptation method, respectively.

During training, as shown in Fig. \ref{fig:ttt}, we select the typical motion in-betweening network \cite{harvey2020robust} as our baseline to generate transition human skeleton motions. The training process is the same as \cite{harvey2020robust}. The training loss is formulated as:
\begin{equation}
L_{train} = L_{inbetween} + \alpha L_{contact},
\end{equation}
where $L_{inbetween}$ is the sum of all the training losses used in \cite{harvey2020robust}, $L_{contact}$ is the foot contact consistency loss, and $\alpha$ is a constant coefficients.

During testing, we only use the self-supervised losses to update the motion in-betweening model for several epochs on the whole test data. The testing loss is formulated as:
\begin{equation}
L_{test} = L_{contact} + \beta L_{smooth},
\end{equation}
where $\beta$ is a constant coefficients. After test time adaptation, the motion in-betweening network is able to produce optimized plausible skeleton motions $\hat{m}_{1:T-1}$. 

\subsection{Skeleton to Humanoid Matching}

\begin{figure}[t]
\centering
\includegraphics[scale=0.32]{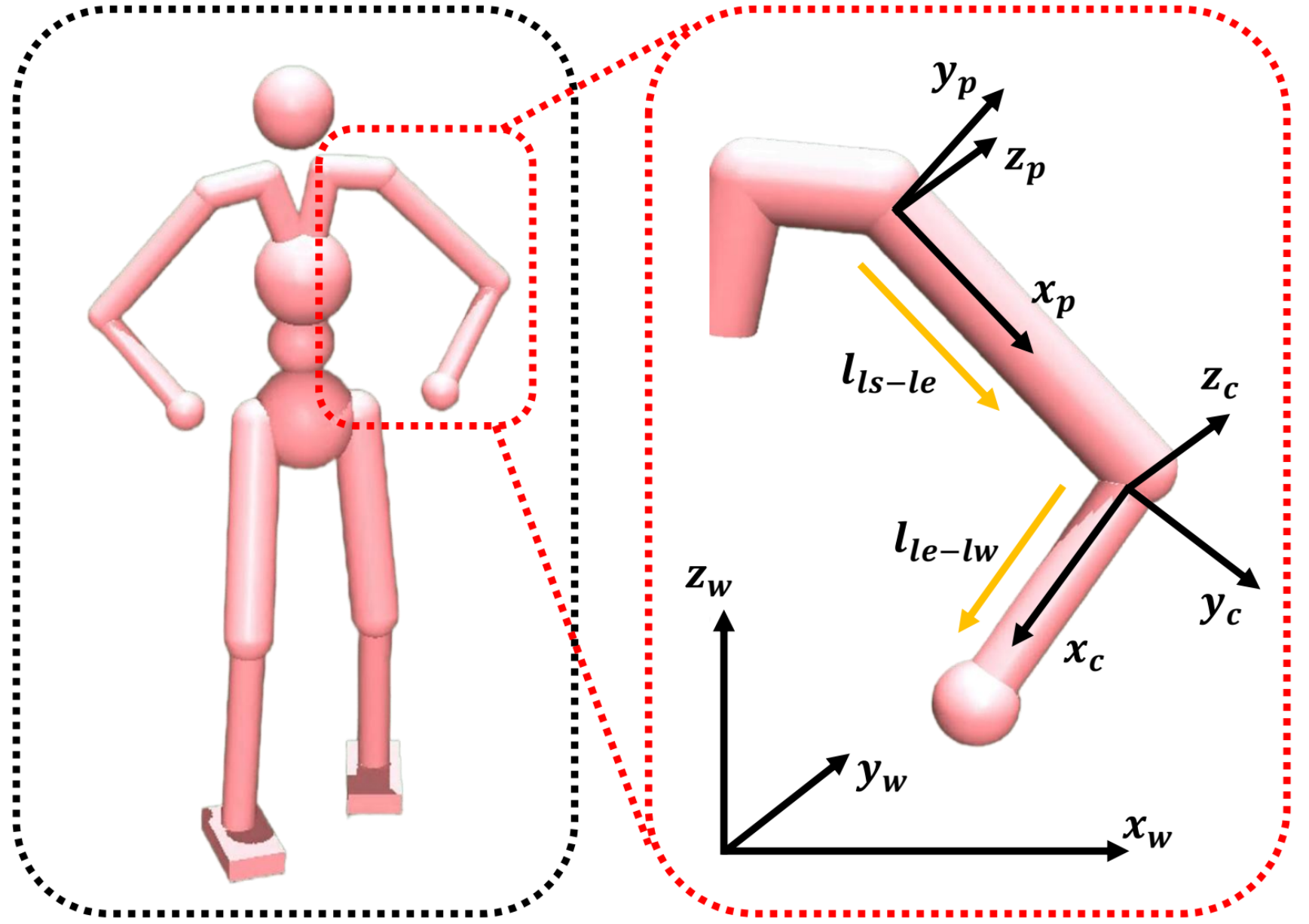}
\vspace{-1mm}
\caption{Illustration of skeleton to humanoid matching process in left elbow. $(x_w,y_w,z_w)$, $(x_c,y_c,z_c)$ and $(x_p,y_p,z_p)$ represent the world coordinate, the child coordinate(left elbow) and the parent coordinate(left shoulder) respectively.}  
\label{fig:ik}       
\end{figure}
\vspace{-1mm}

After acquiring optimized skeleton motions $\hat{m}_{1:T-1}$ from stage I, we then transfer the skeleton motions $\hat{m}_{1:T-1}$ to corresponding humanoid motions $\tilde{m}_{1:T-1}$ which can be imitated by RL. To achieve this, recent works \cite{yuan2021simpoe, luo2021dynamics} constructed a humanoid model which is exactly the same as the skeleton or the SMPL model in shapes and kinematics tree, hence it is easy to transfer the motions to humanoid motions with no error because they can directly transfer the axis-angles of the SMPL model to the corresponding euler angles of humanoid. However, we emphasize that it is not suitable for all cases in practice. 

First, this method is not general and cumbersome because it needs to carefully construct a humanoid model and it is not possible to flexibly transfer motions between human skeleton and humanoid with different shapes and kinematics trees. Second, constructing an exactly same humanoid model is an ideal solution because the pose of humanoid $\tilde{m}_t$ in the physics simulator is represented by euler angles $\tilde{q}_t$ with specific ranges which corresponds body joints constraints. For example, the euler angle of the shoulder joint in z axis is usually between -90 and 90 degrees. However, transferring the quaternions of the skeleton to the corresponding euler angles of humanoid in a specific range is actually an ill-posed problem. The range of the euler angle in each axis is between -180 and 180 degrees, hence one rotation can be represented by many different euler angles. For human skeletons with strange poses or strange kinematics trees, it is difficult to transfer the quaternions to the corresponding euler angles in a specific range.

To solve this problem, inverse kinematics is a solution that can convert the 3d joint positions of skeleton into the euler angles of humanoid. Analytical Inverse kinematics calculates the rotation vectors at the end of a kinematic chain in a given position and can match motions between skeletons and humanoids with different shapes and kinematics trees. Previous works use analytical inverse kinematics \cite{yu2021skeleton2mesh, li2021hybrik} to convert the 3d joint positions of skeleton to the axis-angles of the SMPL model. We extend their works to convert 3d joint positions of skeleton to the euler angles of humanoid via specific matching equations. Concretely, we select a suitable axis definition for each joint of the skeleton that can compute the corresponding euler angle of the humanoid model. We take the left elbow for instance to describe the detailed matching process in Fig. \ref{fig:ik}. Following \cite{yu2021skeleton2mesh}, we consider the skeleton and the humanoid as articulated bodies with pairs of parent-children joints. Suppose the parent and the child coordinate system of joints in 3d skeleton is denoted as $[\mathbf{x}_p, \mathbf{y}_p, \mathbf{z}_p]$ and $[\mathbf{x}_c, \mathbf{y}_c, \mathbf{z}_c]$, then the parent and child coordinate system of the left elbow is calculated by Eqn. \ref{eq: cpdir},

\begin{equation}\label{eq: cpdir}
\left\{ \begin{aligned}
         &\left [ \mathbf{x}_p ,\mathbf{y}_p, \mathbf{z}_p \right ] =\left [ \frac{l_{ls-le} }{\left | l_{ls-le} \right |} , \frac{\mathbf{z}_p\otimes \mathbf{x}_p}{\left | \mathbf{z}_p\otimes \mathbf{x}_p \right | } , \frac{-l_{ls-le} \otimes l_{le-lw} }{\left | l_{ls-le} \otimes l_{le-lw} \right | }  \right ] \\
                  &\left [ \mathbf{x}_c ,\mathbf{y}_c, \mathbf{z}_c \right ] =\left [ \frac{l_{le-lw} }{\left | l_{le-lw} \right |} , \frac{\mathbf{z}_c\otimes \mathbf{x}_c}{\left | \mathbf{z}_c\otimes \mathbf{x}_c \right | } , \mathbf{z}_p  \right ]
        \end{aligned}~, \right.
\end{equation}
where each item (e.g., $\mathbf{x}_c$) is a 3*1 vector in world coordinate system. $\mathbf{I}_{a-b}$ indicates the vector pointing from joint $a$ to joint $b$ and subscripts $ls$, $le$, $lw$ denote left shoulder, left elbow and left wrist, respectively. Specifically, $l_{le-lw}$ means the vector pointing from the left elbow joint to the left wrist joint. Then we can calculate the euler angle $\mathbf{E}$ of the left elbow of the humanoid via Eqn. \ref{dir2euler}, 

\begin{equation}\label{dir2euler}
\left\{ \begin{aligned}
         &\vec{N} = [n_0, n_1, n_2] = [\mathbf{x}_c\cdot \mathbf{x}_p^T, \mathbf{x}_c\cdot \mathbf{y}_p^T, \mathbf{x}_c\cdot \mathbf{z}_p^T]\\
         &\vec{O} = [o_0, o_1, o_2] = [\mathbf{y}_c\cdot \mathbf{x}_p^T, \mathbf{y}_c\cdot \mathbf{y}_p^T, \mathbf{y}_c\cdot \mathbf{z}_p^T]\\
         &\vec{A} = [a_0, a_1, a_2] = [\mathbf{z}_c\cdot \mathbf{x}_p^T, \mathbf{z}_c\cdot \mathbf{y}_p^T, \mathbf{z}_c\cdot \mathbf{z}_p^T]\\
         &\mathbf{E}_z = atan2(\frac{n_1}{n_0})\\
         &\mathbf{E}_y = atan2(\frac{-n_2}{n_0\ast cos(\mathbf{E}_z)+n_1\ast sin(\mathbf{E}_z)} )\\
         &\mathbf{E}_x = atan2(\frac{-a_1\ast cos(\mathbf{E}_z)+a_0\ast sin(\mathbf{E}_z)}{o_1\ast cos(\mathbf{E}_z)-o_0\ast sin(\mathbf{E}_z)} )\\
         &\mathbf{E} = \left[\mathbf{E}_z,\mathbf{E}_y, \mathbf{E}_x \right]\\
        \end{aligned}~, \right.
\end{equation}
where $\mathbf{E}$ is the euler angle of the humanoid model in z-y-x rotation order. After applying the similar matching operation on the other joints of the humanoid, the whole human skeleton motions $\hat{m}_{10:T-1}$ can be converted to humanoid motions $\tilde{m}_{10:T-1}$. In addition, because the matching process is applied frame by frame, the values of the converted euler angles usually flip and become discontinued due to the $atan2$ process in equation \ref{dir2euler}. We correct the values of euler angles by adding a temporally checking process, suppose $\mathbf{E}_z^t$ is a euler angle of a joint in z axis at time t, 
\begin{equation}\label{flipcorrect}
\left\{ \begin{aligned}
         & \mathbf{E}_z^{t+1} = \mathbf{E}_z^{t+1} + 2\pi, \text{ if } \mathbf{E}_z^{t}>0 \text{and}  |\mathbf{E}_z^{t+1}-\mathbf{E}_z^{t}|>\lambda \\
         & \mathbf{E}_z^{t+1} = \mathbf{E}_z^{t+1} - 2\pi, \text{ if } \mathbf{E}_z^{t}<=0 \text{and}  |\mathbf{E}_z^{t+1}-\mathbf{E}_z^{t}|>\lambda \\
        \end{aligned}~, \right.
\end{equation}
where $\lambda$ is a threshold value for checking if the euler angles are discontinued or not and we set $\lambda$ to 5. In the end, we can convert the synthesized skeleton motions $\hat{m}_{1:T-1}$ to humanoid motions $\tilde{m}_{1:T-1}$. 

\begin{figure}[t]
\centering
\includegraphics[scale=0.48]{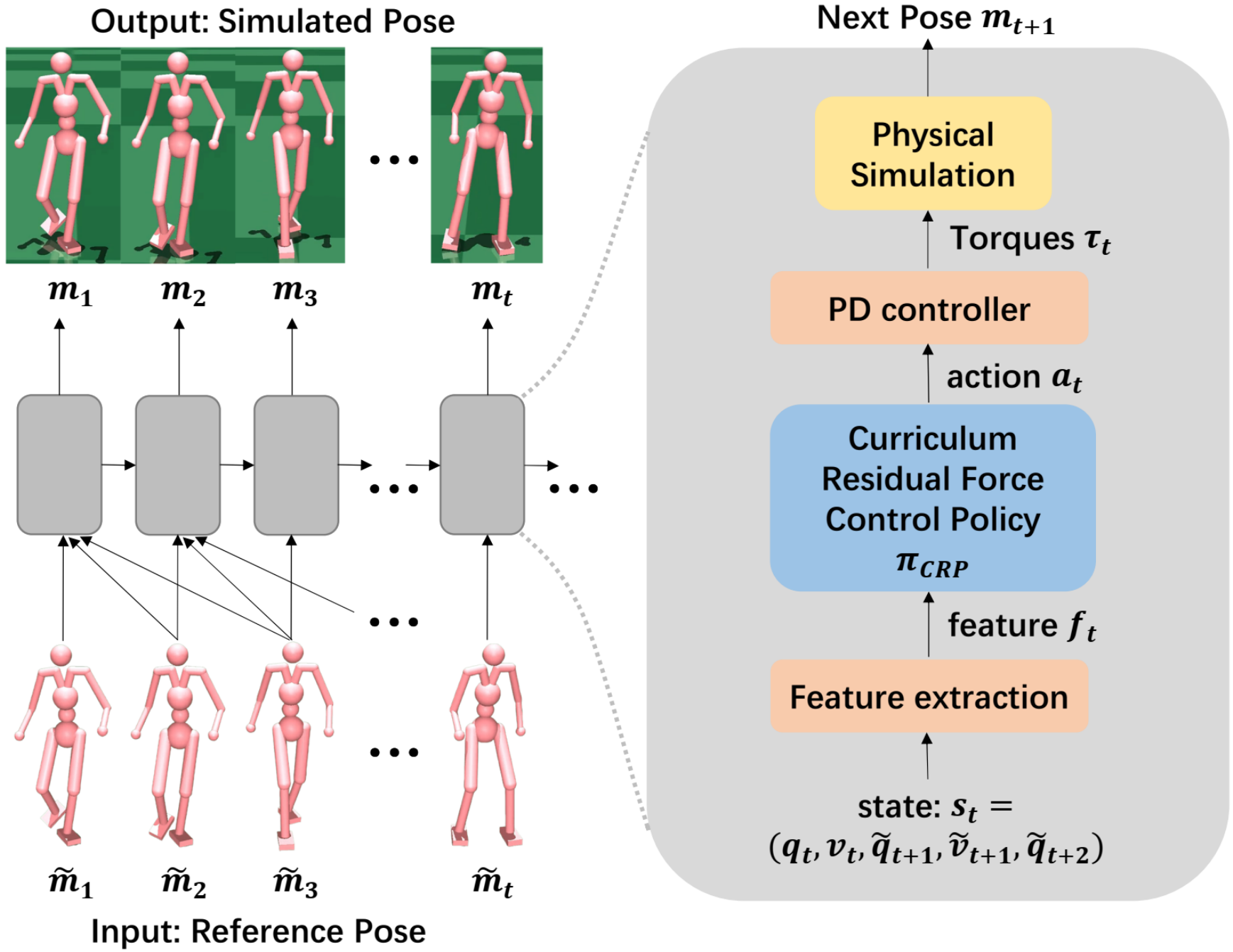}
\caption{Overview of  a curriculum residual force control policy $\mathbf{\pi}_{CRP}$. Our $\mathbf{\pi}_{CRP}$ can actuate the humanoid to iteratively imitate the reference pose in next one frame precisely given the reference pose in next two frames.}
\label{fig:ghp policy}       
\end{figure}

\subsection{Motion Imitation based on RL}

After getting the converted humanoid motions $\tilde{m}_{1:T-1}$, they are still physically implausible due to lack of physical constraint. Motivated by \cite{yuan2020residual, yuan2021simpoe, luo2021dynamics}, we utilize reinforcement learning to learn a curriculum residual force control policy $\pi_{CRP}$ which can imitate $\tilde{m}_{1:T-1}$ to generate physically-plausible humanoid motions $m_{1:T-1}$. The motion imitation problem is usually formulated as a Markov decision process. Given a reference motion and current state $s_t$, the humanoid agent interacts with the physics simulator environment by action $a_t$ and receive reward $r_t$. The action is produced by a policy $\pi_{CRP}(a_t|s_t)$ conditioned on state $s_t$. The reward encourages the agent to act like the reference motion. When an action is taken, the current agent state $s_t$ changes to next state $s_{t+1}$. The goal of the motion imitation is to learn a policy that can maximizes the average cumulative rewards.

\subsubsection{Motion Imitation Paradigm} 
Here we detail the states, actions, rewards and the training process.

\noindent\textbf{State.} Previous work \cite{yuan2021simpoe,peng2018deepmimic,yuan2020residual} construct the humanoid state $s_t$ which contains the information of the character's current pose and the information of the reference pose in the next frame. We utilize the information of the reference pose in the next two frames that can learn a finer control policy. Specifically, the state $s_t\triangleq(q_t, v_t, \tilde{q}_{t+1}, \tilde{v}_{t+1}, \tilde{q}_{t+2})$ and we apply a hand-craft feature extraction to obtain informative features $f_t$ as the input of $\pi_{CRP}$ from $s_t$. We formulate $f_t\triangleq(q_t, v_t, j_{t}, g_{t}, (\tilde{q}_{t+1}-q_t), (\tilde{j}_{t+1} - j_{t}), (\tilde{g}_{t+1} - g_{t}), (\tilde{v}_{t+1}-v_t), (\tilde{q}_{t+2}-q_{t}), (\tilde{j}_{t+2} - j_{t}), (\tilde{g}_{t+2} - g_{t}), (\tilde{v}_{t+2}-v_{t}))$, where $q_t$ is the euler angle of the simulated humanoid, $v_t$ is the joint velocity computed by finite difference, $j_{t}$ and $g_{t}$ are the 3d joint position in local head coordinates and global coordinates, $\tilde{q}_{t+1}$ is the euler angle of reference pose in time t+1, $\tilde{v}_{t+1}$ the joint velocity of the reference pose in time t+1, $\tilde{j}_{t+1}$ and $\tilde{g}_{t+1}$ are the 3d joint position of the reference pose in local head coordinates and global coordinates.

\noindent\textbf{Action.} 
Following prior works \cite{yuan2021simpoe, luo2021dynamics,yuan2020residual}, we utilize proportional derivative (PD) controller \cite{tan2011stable} at each non-root joint to generate torques. Under this setting, the action $a_t$ is the target joint angles $u_t$ of the PD controllers. The joint torques $\tau_t$ are specifically calculated by:
\begin{equation}
\tau_t = k_p \circ (u_t - q_t) - k_d \circ \dot{q}_t,
\end{equation}
where $k_p$ and $k_d$ are manually defined parameters and $\circ$ is element-wise multiplication. The PD controllers act like damped springs that drive joints to target angles $u_t$, while $k_p$ and $k_d$ are the stiffness and damping of the springs. Prior works \cite{yuan2020residual,luo2021dynamics} also allowed the policy to apply external residual forces $\eta_t = k_r \circ S_r$ to joints to improve the motion imitation performance, $S_r$ is a predefined residual force scale and $k_r \in[0,1]$ is a weighting parameter learned by policy . In section \ref{crf}, we will introduce our curriculum residual force learning that can dynamically adjust the residual force scale $S_r$ in RL training. Overall, our action is $a_t\triangleq(u_t, \eta_t)$.

\noindent\textbf{Reward.} Following \cite{luo2021dynamics,yuan2020residual}, we implement the same reward function that encourages the policy to mimic the reference pose.

\subsubsection{Curriculum Learning for Residual Forces}\label{crf}
The residual forces $\tau_t$ proposed by \cite{yuan2020residual} aims at utilizing additional forces to enable finer control of the humanoid character. Recent works \cite{yuan2020residual,yuan2021simpoe} used it to perform humanoid motions with extra interaction such sitting on the chair even there is no chair in the physics simulator. However, large scale residual forces will make humanoids perform some unnatural motions or overly energetic movements. Meanwhile, removing the residual forces will cause the low converge speed of RL and a bad imitation performance. To balance the trade-off between the motion naturalness and the imitation performance, we propose a curriculum learning strategy that gradually adjusts the residual force scale in the training of reinforcement learning.

Starting from a large residual forces, we gradually decrease the residual force scale in RL training. Suppose $i$ is the training iteration of PPO RL algorithm, $S_r$ is residual force scale which is adjusted by:
\begin{equation}
S_{r}  = \begin{cases}
  & S^{init}_{r}, \text{ if } 0\leq i \leq i_{start} \\
  & S^{init}_{r} - \alpha*(i-i_{start}), \text{ if } i_{start}\leq i \leq i_{end} \\
  & S^{init}_{r} - \alpha*(i_{end}-i_{start}), \text{ if } i\geq i_{end}
\end{cases},
\end{equation}
where $i_{start}$ and $i_{end}$ are the start and end iterations of the curriculum, and $\alpha$ is a hyper-parameter indicating the changing rate of $S_r$. We empirically set $S^{init}_{r}=220, i_{end} = 1300, i_{start} = 100, \alpha = 0.1$ and train RL for 1500 iterations. By introducing curriculum learning for residual forces, the policy can quickly learn to imitate the reference motion precisely with the help of large residual forces and learn to maintain the imitation performance while getting rid of large residual forces.

\begin{figure*}[t]
\centering
\includegraphics[scale=1]{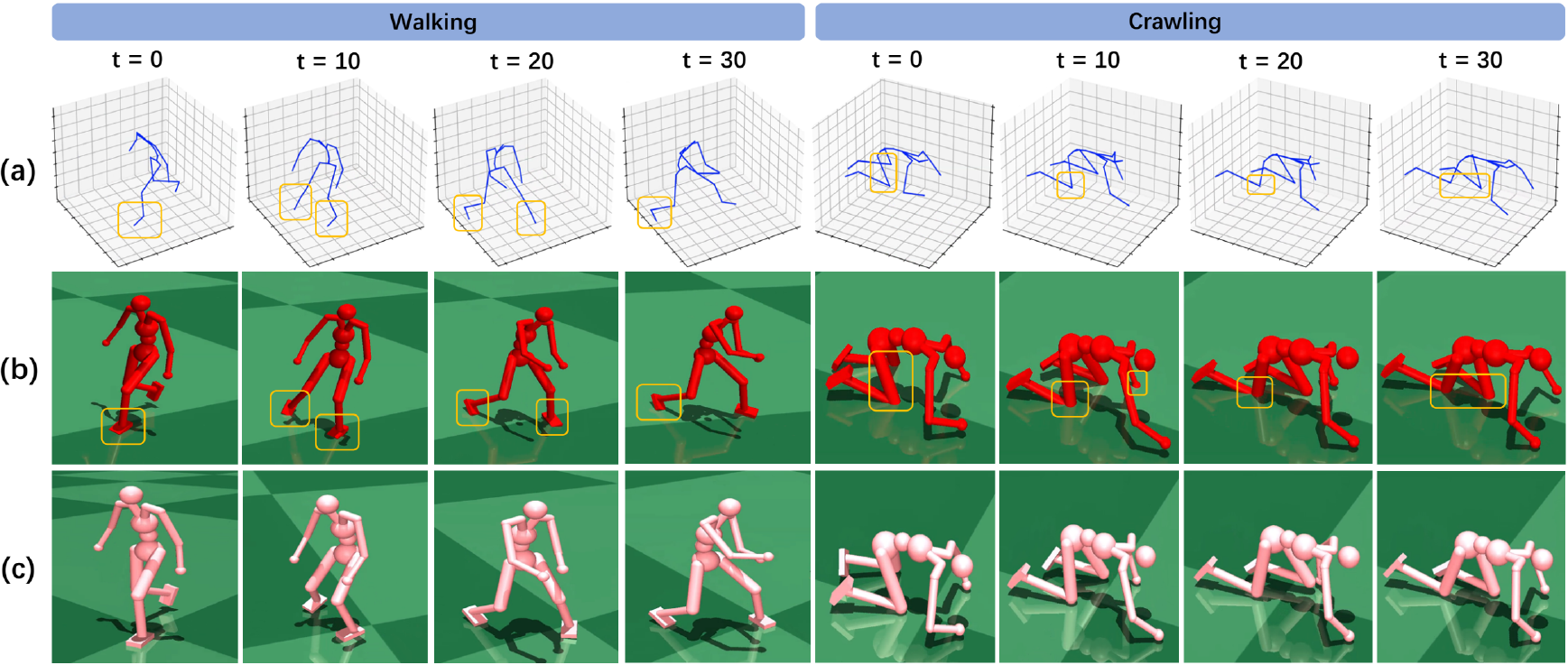}
\vspace{-2mm}
\caption{Qualitative Results of motions synthesized by our Skeleton2Humanoid system. The top row shows the skeleton motions ({\color{blue}blue}) after our test time adaptation stage (TTA), the middle row shows the matched humanoid motions ({\color{red}red}) by our analytical inverse kinematics method (IK), the bottom row shows our final imitated humanoid motions ({\color{pink}pink}) by our curriculum residual force control policy $\pi_{CRP}$. Ground penetration is highlighted with boxes.}
\label{qualitative results}       
\end{figure*}

\begin{table*}[]
 \resizebox{\textwidth}{!}{
\begin{tabular}{llllllllllllllll}
&   & FP &                         &   & FQ &                         &   & JQ &                         &   & SM &   &   & L2P & \\ \cline{2-16} 
length              & 5 & 15  & \multicolumn{1}{l|}{30} & 5 & 15 & \multicolumn{1}{l|}{30} & 5 & 15 & \multicolumn{1}{l|}{30} & 5 & 15 & \multicolumn{1}{l|}{30} & 5 & 15 & 30 \\ \hline
baseline\cite{harvey2020robust}   &  -0.47 &  -0.43  & \multicolumn{1}{l|}{-1.43}   &1.04\%   &  4.12\%  & \multicolumn{1}{l|}{8.37\%}   & 0.68\%  &  1.25\%  & \multicolumn{1}{l|}{2.08\%}   & 20.8  &  20.8  & \multicolumn{1}{l|}{20.2}  & 0.19 &  0.55    & {1.16}\\
baseline w/ TTA    & \bf{-0.04}  & -0.29   & \multicolumn{1}{l|}{-1.13}   & \bf{0.89\%}  &  2.91\%  & \multicolumn{1}{l|}{6.57\%}   & \bf{0.07\%}  & 1.08\%   & \multicolumn{1}{l|}{1.79\%}   & 20.6  &  20.6  & \multicolumn{1}{l|}{19.9}   & \bf{0.16}  &  \bf{0.53}   & {\bf{1.13}}\\
baseline w/ TTA,IK  & -0.24  &  -0.63  & \multicolumn{1}{l|}{-1.73}   & 6.28\%  &  7.70\%  & \multicolumn{1}{l|}{11.3\%}   & 1.49\%  & 1.80\%   & \multicolumn{1}{l|}{2.53\%}   & 20.6  &  20.6  &  \multicolumn{1}{l|}{20.0}  & {0.23}  &  0.56   & {1.16} \\

baseline w/ TTA,IK,CRP  & -0.06  & \bf{-0.04}   & \multicolumn{1}{l|}{\bf{-0.08}}   & 1.48\%  &  \bf{0.73\%}  & \multicolumn{1}{l|}{\bf{0.80\%}}   & 0.35\%  &  \bf{0.16\%}  & \multicolumn{1}{l|}{\bf{0.16\%}}   & \bf{20.5}  &  \bf{20.5}  &   \multicolumn{1}{l|}{\bf{19.8}} & 0.34  &  0.68   & {1.27}
\\ \hline
\end{tabular}}
\caption{Evaluation of our Skeleton2Humanoid \emph{i.e.} baseline w/ TTA,IK,CRP on Noobjects LaFAN1 dataset. "TTA", "IK" and "CRP" represent the test time adaptation, analytical inverse kinematics and curriculum residual force control policy respectively.}
\label{inbetweening}
\vspace{-2mm}
\end{table*}
\vspace{-2mm}

\section{Experiments}
In this section, we perform experimental evaluations of our Skeleton2Humanoid system on the motion realism and accuracy, then we analyze the properties of the test time adaptation stage, the skeleton to humanoid matching stage and the motion imitation stage based on RL individually.

\noindent\textbf{Dataset}. To demonstrate the effectiveness of our Skeleton2Humanoid system on the motion in-betweening task, We employ the following datasets in our experiments: (1) \textbf{LaFAN1} \cite{harvey2020robust}, which is a widely used large scale dataset for evaluating motion in-betweening performance. LaFAN1 dataset contains various complex motions such as fighting, crawling and climbing obstacles. To be noticed, We only evaluate the test time adaptation stage of our Skeleton2Humanoid system on LaFAN1 dataset and we will explain it later. (2) \textbf{Noobjects LaFAN1} is a subset of LaFAN1 dataset for evaluating the physical plausibility of our synthesized in-betweening motions. LaFAN1 dataset contains many human object interaction motions such as climbing obstacles and we can not construct objects in the physics simulator due to the absence of the ground truth of 3d objects, hence we remove these motions and only consider human motions performing on the ground. We evaluate our whole Skeleton2Humanoid system on this dataset.

\noindent\textbf{Evaluation Metrics.} We use the following metrics including physics-based and accuracy-based metrics for all following evaluations: (1) \textbf{FP}, which is the foot penetration that computes the mean penetration distance of feet joints and ground following \cite{rempe2021humor}.  (2) \textbf{FQ}, which computes the frequency of the penetrations between the feet and ground penetrations. (3) \textbf{JQ}, which similarly computes the frequency of penetrations between all the joints and the ground. (4) \textbf{SM}, which represents motion smoothness that computes the average L2 distance between body joints within two adjacent frames. (5) \textbf{MPJPE}, which is the mean per joint position error for evaluating the mismatch between poses. (6) \textbf{L2Q} and \textbf{L2P}, where L2Q denotes the the average L2 distances of the global quaternions between the predicted motions and their ground truth. Similarly, L2P denotes the average L2 distance of the global positions. We follow standard evaluations from  \cite{harvey2020robust} to compute them.

\vspace{-2mm}
\subsection{Evaluation of Skeleton2humanoid System}
\noindent\textbf{Experiment Setting}. We evaluate our whole Skeleton2Humanoid system on the Noobjects LaFAN1 dataset. We select the typical state-of-the-art motion in-betweening approach Harvey et al. \cite{harvey2020robust} as a baseline and validate our approach based on it. The baseline method is trained with various transition lengths (5 frames to 30 frames) between past keyframes and a future keyframe and is evaluated when transition lengths are 5 frames, 15 frames and 30 frames.

\noindent\textbf{Results and Discussion}. 
The results on the Noobjects LaFAN1 dataset are summarized in table \ref{inbetweening}. First, it is evident that our Skeleton2Humanoid outperforms the baseline in all physics-based metrics and achieves comparable performance in accuracy-based metrics. In particular, Skeleton2Humanoid achieve significantly lower FP, FQ, JQ and SM, which demonstrates that utilizing the simulated humanoid character can effectively reduce the physical artifacts such as the penetration problem between human bodies and the ground and can produce physically-plausible motions. Second, we can observe that our test time adaptation stage (TTA) outperforms the baseline in both physics-based metrics and accuracy-based metrics, demonstrating that our TTA can optimize the motion in-betweening model to produce not only plausible but also accurate motions. Third, we find that applying the analytical inverse kinematics and the curriculum residual force control policy (CRP) cause worse L2P. The reason is that our efficient analytical inverse kinematics strategy can not perfectly matching the skeleton and humanoid even its matching error is already really low (MPJPE: 11.29mm) as discussed in section. \ref{matching exp}, and our state-of-the-art motion imitation policy CRP as shown in section. \ref{policy exp} still struggle to imitate various complex in-betweening motions perfectly. Such a phenomenon is also observed in \cite{luo2021dynamics} when they utilize RL to imitate human motions for ego pose estimation. We emphasize that current motion in-betweening method should not only consider the accuracy of synthesized motions but also consider the physical plausibility.

We also provide qualitative results in Fig. \ref{qualitative results}, we can see that the imitated humanoid motions can accurately synthesize joint contact with the ground and without penetration. Our state-of-the-art RL policy is also able to control the humanoid to perform complex motions such as crawling. We provide more qualitative results in supplementary material for demonstrating the physical plausibility of our synthesized humanoid motions.

\subsection{Evaluation of Each Stage}
In this section, we concretely analyze the effect of each stage in our Skeleton2Humanoid system.

\subsubsection{Evaluation of Test Time Motion In-betweening Network Adaptation Stage} 

\begin{table*}
\begin{tabular}{lllllllllllll}
         &   & FP &                         &   & FQ &                         &   & JQ &                         &   & SM &    \\ \cline{2-13} 
length   & 5 & 15 & \multicolumn{1}{l|}{30} & 5 & 15 & \multicolumn{1}{l|}{30} & 5 & 15 & \multicolumn{1}{l|}{30} & 5 & 15 & 30 \\ \hline
baseline \cite{harvey2020robust}* & -0.039  & -0.351   & \multicolumn{1}{l|}{-1.186} & 0.884\%  & 3.49\%   & \multicolumn{1}{l|}{7.04\%}   & 0.538\%  & 1.02\%   & \multicolumn{1}{l|}{1.71\%}   & 20.7  & 20.9   &  20.6 \\
baseline w/ TTA  & \bf{-0.036}  & \bf{-0.235}   & \multicolumn{1}{l|}{\bf{-0.930}}  & \bf{0.775\%}  & \bf{2.42\%}   & \multicolumn{1}{l|}{\bf{5.52\%}} & \bf{0.527\%}  & \bf{0.861\%}   & \multicolumn{1}{l|}{\bf{1.46\%} }  & \bf{20.5}  & \bf{20.5}   &  \bf{20.0}  \\
\hline
\end{tabular}
\caption{Evaluation of test time adaptation stage in terms of physics-based metrics on LaFAN1 dataset. "*" means self-implementation (better results than the original paper). "TTA" represents the test time adaptation.}
\label{ttt_physics}
\vspace{-2mm}
\end{table*}
\vspace{-2mm}

\begin{table}
\scalebox{0.9}{
\begin{tabular}{lllllll}
         &   & L2Q &                         &   & L2P &    \\ \cline{2-7} 
length   & 5 & 15  & \multicolumn{1}{l|}{30} & 5 & 15  & 30 \\ \hline
Zero-Vel & 0.56  &  1.10   & \multicolumn{1}{l|}{1.51}   & 1.52  & 3.69    & 6.60    \\
Interp   & 0.22  &  0.62   & \multicolumn{1}{l|}{0.98}   &  0.37 & 1.25    & 2.32    \\ \hline
Kaufmann et al. \cite{kaufmann2020convolutional}  & 0.49  & 0.60    & \multicolumn{1}{l|}{0.78}   & 0.84  & 1.07 & 1.53   \\
duan et al. \cite{duan2022unified} & 0.18  &  0.37 & \multicolumn{1}{l|}{\bf{0.61}} & 0.23  & 0.56  & \bf{1.06}  \\
baseline \cite{harvey2020robust} & 0.17  & 0.42    &  \multicolumn{1}{l|}{0.69} & 0.23  &  0.65   &  1.28  \\
baseline \cite{harvey2020robust}* &  0.138 & 0.373 & \multicolumn{1}{l|}{0.667} & 0.183  & 0.548  & 1.167   \\
baseline w/ ablation    & 0.134 & 0.367  &  \multicolumn{1}{l|}{0.662} & 0.162  &  0.527 &1.142    \\
baseline w/ TTA  &  \bf{0.133} & \bf{0.367} &  \multicolumn{1}{l|}{0.661}  & \bf{0.162}  &  \bf{0.526} &   1.140 \\ \hline
\end{tabular}}
\caption{Evaluation of test time adaptation stage in terms of accuracy-based metrics on LaFAN1 dataset. "TTA" represents our test time adaptation. "ablation" represents applying test time adaptation without motion smoothness loss.}
\label{ttt_acc}
\vspace{-2mm}
\end{table}
\vspace{-2mm}

 We evaluate our test time adaptation stage on the test set of the whole LaFAN1 dataset \cite{harvey2020robust}. Since no domain generalization approach or test time adaptation approach has been proposed for motion synthesis tasks, we compare against the state-of-the-art motion in-betweening approaches including Kaufmann et al. \cite{kaufmann2020convolutional}, duan et al. \cite{duan2022unified} and our implemented baseline Harvey et al. \cite{harvey2020robust}. We select L2Q, L2P, FP, FQ, JQ, SM as evaluation metrics.


\noindent\textbf{Results and discussion}
The results are shown in Table \ref{ttt_acc} and Table \ref{ttt_physics}. First, As shown in Table \ref{ttt_physics}, we observe that our test time adaptation stage (TTA) outperforms the baseline approach regarding all physics-based metrics. It shows that TTA can help reduce the physical artifacts and produce more physically-plausible motions for the humanoid character to imitate. Second, we can see that with our test time adaptation stage, our adapted motion in-betweening network significantly outperforms comparison approaches regarding all accuracy-based metrics in Table \ref{ttt_acc}. We also achieve a new state-of-the-art performance in the L2Q and the L2P metrics. The ablation (w/o ablation, i.e. w/o smooth loss) shows that the motion smooth loss is also useful for test time adaptation.

\subsubsection{Evaluation of Skeleton to Humanoid Matching Stage}\label{matching exp}
To evaluate our analytical inverse kinematics (IK) method, we report the matching performance on the synthesized in-betweening motions (transition lengths: 30 frames) by TTA on the test set of Noobject LaFAN1 dataset by computing the mean
per joint position error (MPJPE) and inference time. Our analytical inverse kinematics method achieves excellent performance in terms of precision (MPJPE: 11.29mm) and speed (0.122s/sequence). It is worth noting that even our method precisely transfers the skeleton motions to humanoid motions, it still causes a slight performance degradation problem in the L2P metric as shown in table \ref{inbetweening} (baseline w/ TTA, IK). This shows that current state-of-the-art motion in-between methods can already predict transition motions with very high accuracy, hence a slight difference in poses will sensitively influence the prediction accuracy.

\subsubsection{Evaluation of Motion Imitation based on RL Stage}\label{policy exp}
Now we detailedly study the performance of our proposed curriculum residual force control policy $\pi_{CRP}$, we evaluate the motion imitation performance on the synthesized humanoid motions with transition lengths of 30 frames by our test time adaptation and skeleton to humanoid matching on the test set of Noobject LaFAN1 dataset. For baseline method, we select universal humanoid controller (UHC) \cite{luo2021dynamics} which is a state-of-the-art motion imitation approach based on residual force control \cite{yuan2020residual} for comparison. Since our framework uses different humanoid characters, we use an in-house implementation of UHC \cite{luo2021dynamics}. We adopt the same residual force scale and state feature in each frame for fair comparison. All the policies are trained in 1500 iterations and we report the L2P and the MPJPE metrics.

\begin{table}
\begin{center}
\setlength{\tabcolsep}{2.0mm}{
\begin{tabular}{c|c|c}
\hline
Method & L2P ($\downarrow$) & MPJPE ($\downarrow$)  \\
\hline\hline
 UHC \cite{yuan2021simpoe}  & 1.307 & 31.68mm \\
 CRP w/o TwoFrameFeats & 1.281 & 29.16mm \\
 CRP w/o CurForce & 1.292 & 29.95mm \\
 CRP (Ours)  & \bf{1.272} & \bf{27.63}mm \\
\hline
\end{tabular}}
\end{center}
\caption{Evaluation of our curriculum residual force control policy $\pi_{CRP}$ on Noobjects LaFAN1 dataset. "CurForce" represents applying curriculum learning for residual forces. "TwoFrameFeats" represents utilizing state information from target poses of next two frames}
\label{uhp}
\vspace{-3mm}
\end{table}

\begin{figure}[t]
\centering
\includegraphics[scale=0.31]{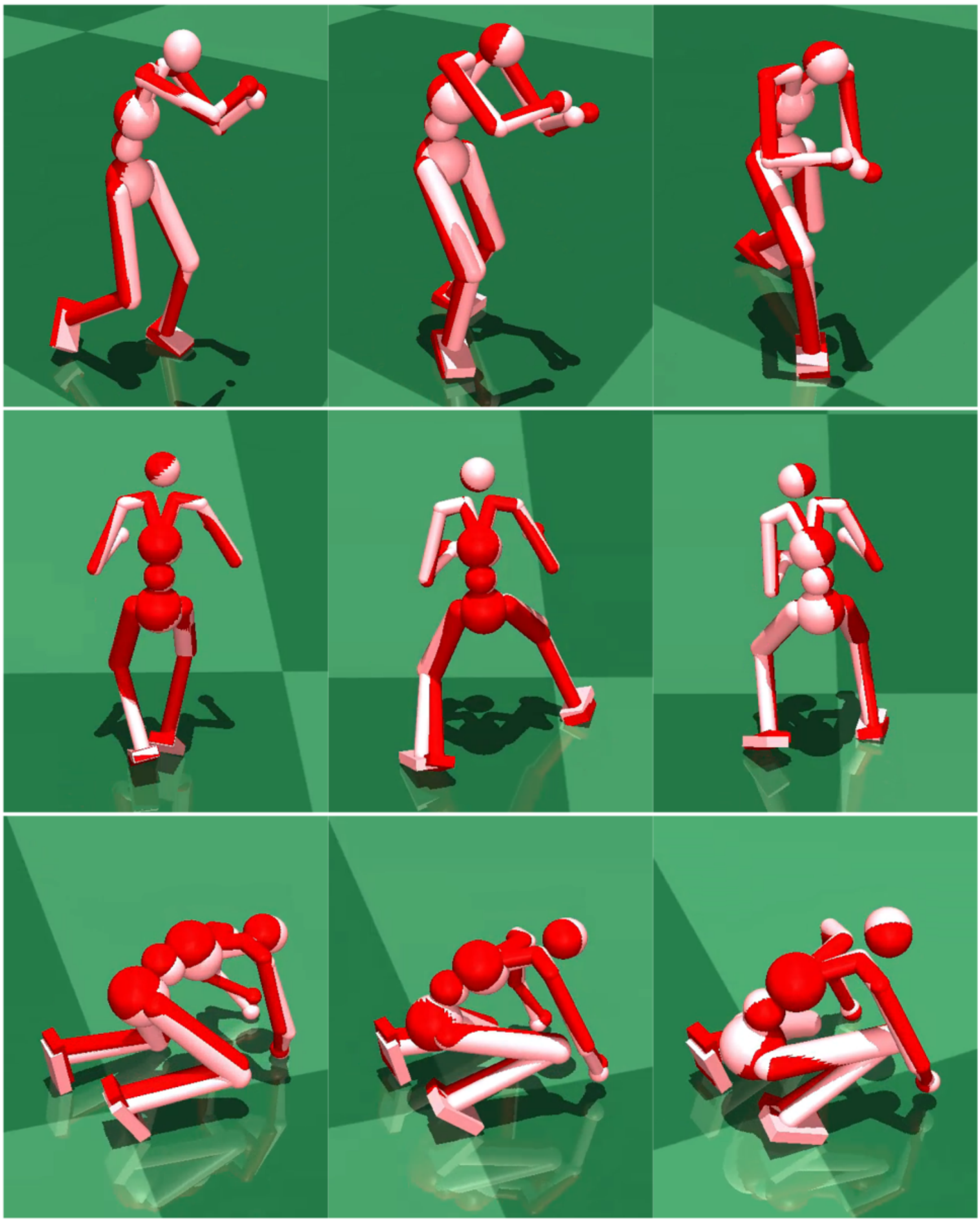}
\vspace{-2mm}
\caption{Visualization of the reference motions ({\color{red}red humanoids}) and imitated motions ({\color{pink}pink humanoids}) by our curriculum residual force control policy $\pi_{CRP}$.}
\label{imitate results}       
\end{figure}

The results are shown in Table \ref{uhp}. First, we can see that our $\pi_{CRP}$ can imitate the synthesized motion with high accuracy, and significantly outperform the baseline approach. Second, from the ablation study, we can observe that the curriculum learning of residual force contributes to better motion imitation accuracy as shown by the corresponding ablations (w/o CurForce and w/o TwoFrameFeats). With the curriculum learning of residual force, the policy can easily learn a optimal control policy with large residual force and gradually learn to control the character precisely while getting rid of the large residual force. We also observe that utilizing the feature of reference poses in next two frames is instrumental, because it extracts informative features from more frames to learn control that advances the character to the next pose.

We also provide qualitative results in Fig. \ref{imitate results}, we can see that our curriculum residual force control policy can imitate many complex motions accurately. More qualitative results can be seen in supplementary material.

\section{Conclusion}
In this paper, we proposed a system ``Skeleton2Humanoid'' which performs physics-oriented motion correction at test time by regularizing synthesized skeleton motions in a physics simulator. Concretely, our system consists of three sequential stages: (I) test time motion synthesis network adaptation, (II) skeleton to humanoid matching and (III) motion imitation based on reinforcement learning (RL). We verified our system on the motion-in-betweening task and showed our method can significantly improve the physical plausibility of synthesized motions. For future work, we plan to transfer the system to other tasks and support more complex human motions including human-object interactions or human-human interactions.

\begin{acks}
This work was supported by NSFC 62176159 and 62101326, China Postdoctoral Science Foundation (2022M712090), Natural Science Foundation of Shanghai 21ZR1432200 and Shanghai Municipal Science and Technology Major Project 2021SHZDZX0102.
\end{acks}


\bibliographystyle{ACM-Reference-Format}
\bibliography{sample-sigconf}

\appendix

\begin{figure*}[t]
  \centering
  \includegraphics[scale = 0.85]{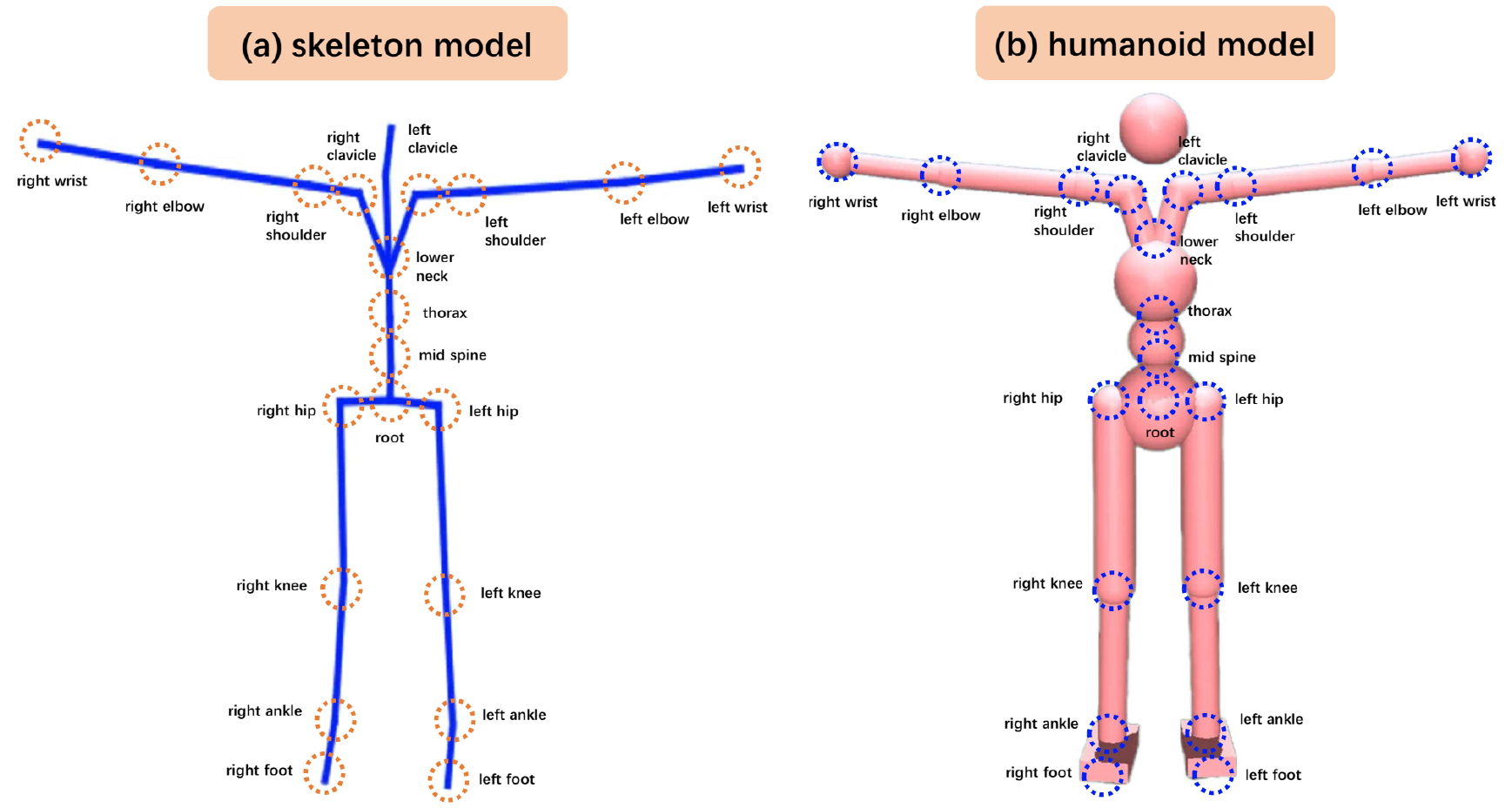}
  \caption{Visualization of the skeleton model (a) and the humanoid model (b). Each model contains 20 joints.}
  \label{fig:supp_ik}
\end{figure*}

\section{Analytical Inverse Kinematics}
Here we concretely introduce our analytical inverse kinematics approach used in skeleton to humanoid matching stage, which is able to convert human skeleton motions to humanoid motions even when the body structure is different from the human skeleton.

We first construct a humanoid model whose body structure (bone length, joint number) is identical to the skeleton except for the kinematics tree. Concretely, the bone length and the joint number of the humanoid and the skeleton are the same, but the initial pose of the humanoid and the skeleton are different and each joint of the humanoid and the skeleton rotates in a different order. We visualize the humanoid model and the skeleton in Fig. \ref{fig:supp_ik}, each model contains 20 joints, and we only calculate the rotations of 15 joints from the positions of all 20 joints. We describe the detailed matching process for 15 calculated joints:

(1) left elbow:
\begin{equation}\label{eq: cpdir}
\left\{ \begin{aligned}
         &\left [ \mathbf{x}_p ,\mathbf{y}_p, \mathbf{z}_p \right ] =\left [ \frac{l_{ls-le} }{\left | l_{ls-le} \right |} , \frac{\mathbf{z}_p\otimes \mathbf{x}_p}{\left | \mathbf{z}_p\otimes \mathbf{x}_p \right | } , \frac{-l_{ls-le} \otimes l_{le-lw} }{\left | l_{ls-le} \otimes l_{le-lw} \right | }  \right ] \\
                  &\left [ \mathbf{x}_c ,\mathbf{y}_c, \mathbf{z}_c \right ] =\left [ \frac{l_{le-lw} }{\left | l_{le-lw} \right |} , \frac{\mathbf{z}_c\otimes \mathbf{x}_c}{\left | \mathbf{z}_c\otimes \mathbf{x}_c \right | } , \mathbf{z}_p  \right ]
        \end{aligned}~, \right.
\end{equation}

(2) left shoulder:
\begin{equation}\label{eq: cpdir}
\left\{ \begin{aligned}
&\left [ \mathbf{x}_c ,\mathbf{y}_c, \mathbf{z}_c \right ] =\left [ \frac{l_{ls-le} }{\left | l_{ls-le} \right |} , \frac{\mathbf{z}_c\otimes \mathbf{x}_c}{\left | \mathbf{z}_c\otimes \mathbf{x}_c \right | } , \frac{-l_{ls-le} \otimes l_{le-lw} }{\left | l_{ls-le} \otimes l_{le-lw} \right | }  \right ]\\
         &\left [ \mathbf{x}_p ,\mathbf{y}_p, \mathbf{z}_p \right ] =\left [ \frac{l_{lcl-ls} }{\left | l_{lcl-ls} \right |} , \frac{\mathbf{z}_p\otimes \mathbf{x}_p}{\left | \mathbf{z}_p\otimes \mathbf{x}_p \right | } , \frac{l_{lcl-mspine} \otimes \mathbf{x}_p}{\left | l_{lcl-mspine} \otimes \mathbf{x}_p\mathbf{x}_p \right | }  \right ]         \end{aligned}~, \right.
\end{equation}

(3) left clavicle:
\begin{equation}\label{eq: cpdir}
\left\{ \begin{aligned}
&\left [ \mathbf{x}_c ,\mathbf{y}_c, \mathbf{z}_c \right ] =\left [ \frac{l_{lcl-ls} }{\left | l_{lcl-ls} \right |} , \frac{\mathbf{z}_c\otimes \mathbf{x}_c}{\left | \mathbf{z}_c\otimes \mathbf{x}_c \right | } , \frac{l_{lcl-mspine} \otimes \mathbf{x}_c}{\left | l_{lcl-mspine} \otimes \mathbf{x}_c\mathbf{x}_c \right | }  \right ] \\
         &\left [ \mathbf{x}_p ,\mathbf{y}_p, \mathbf{z}_p \right ] =\left [ \frac{l_{rcl-lcl} }{\left | l_{rcl-lcl} \right |} , \frac{\mathbf{z}_p\otimes \mathbf{x}_p}{\left | \mathbf{z}_p\otimes \mathbf{x}_p \right | } , \frac{l_{lownec-lcl} \otimes l_{lownec-rcl} }{\left | l_{lownec-lcl} \otimes l_{lownec-rcl} \right | }  \right ] 
        \end{aligned}~, \right.
\end{equation}

(4) right elbow:
\begin{equation}\label{eq: cpdir}
\left\{ \begin{aligned}
                  &\left [ \mathbf{x}_c ,\mathbf{y}_c, \mathbf{z}_c \right ] =\left [ \frac{l_{rw-re} }{\left | l_{rw-re} \right |} , \frac{\mathbf{z}_c\otimes \mathbf{x}_c}{\left | \mathbf{z}_c\otimes \mathbf{x}_c \right | } ,
                  \frac{l_{re-rw}\otimes l_{re-rs}}{|l_{re-rw}\otimes l_{re-rs}|}  \right ] \\
                           &\left [ \mathbf{x}_p ,\mathbf{y}_p, \mathbf{z}_p \right ] =\left [ \frac{l_{re-rs} }{\left | l_{re-rs} \right |} , 
                           \frac{\mathbf{z}_p\otimes \mathbf{x}_p}{\left | \mathbf{z}_p\otimes \mathbf{x}_p \right | } , \mathbf{z}_c  \right ] 
        \end{aligned}~, \right.
\end{equation}

(5) right shoulder: 
\begin{equation}\label{eq: cpdir}
\left\{ \begin{aligned}
&\left [ \mathbf{x}_c ,\mathbf{y}_c, \mathbf{z}_c \right ] =\left [ \frac{l_{re-rs} }{\left | l_{re-rs} \right |} , \frac{\mathbf{z}_c\otimes \mathbf{x}_c}{\left | \mathbf{z}_c\otimes \mathbf{x}_c \right | } , \frac{l_{re-rw}\otimes l_{re-rs}}{|l_{re-rw}\otimes l_{re-rs}|} \right ]\\
         &\left [ \mathbf{x}_p ,\mathbf{y}_p, \mathbf{z}_p \right ] =\left [ \frac{l_{rs-rcl} }{\left | l_{rs-rcl} \right |} , \frac{\mathbf{z}_p\otimes \mathbf{x}_p}{\left | \mathbf{z}_p\otimes \mathbf{x}_p \right | } , \frac{l_{rcl-mspine} \otimes \mathbf{x}_p}{\left | l_{rcl-mspine} \otimes \mathbf{x}_p \right | }  \right ] 
        \end{aligned}~, \right.
\end{equation}

(6) right clavicle:
\begin{equation}\label{eq: cpdir}
\left\{ \begin{aligned}
&\left [ \mathbf{x}_c ,\mathbf{y}_c, \mathbf{z}_c \right ] =\left [ \frac{l_{rs-rcl} }{\left | l_{rs-rcl} \right |} , \frac{\mathbf{z}_c\otimes \mathbf{x}_c}{\left | \mathbf{z}_c\otimes \mathbf{x}_c \right | } , \frac{l_{rcl-mspine} \otimes \mathbf{x}_c}{\left | l_{rcl-mspine} \otimes \mathbf{x}_c \right | }  \right ] \\
         &\left [ \mathbf{x}_p ,\mathbf{y}_p, \mathbf{z}_p \right ] =\left [ \frac{l_{rcl-lcl} }{\left | l_{rcl-lcl} \right |} , \frac{\mathbf{z}_p\otimes \mathbf{x}_p}{\left | \mathbf{z}_p\otimes \mathbf{x}_p \right | } , \frac{l_{lownec-lcl} \otimes l_{lownec-rcl} }{\left | l_{lownec-lcl} \otimes l_{lownec-rcl} \right | }  \right ] 
        \end{aligned}~, \right.
\end{equation}

(7) lower neck
\begin{equation}\label{eq: cpdir}
\left\{ \begin{aligned}
&\left [ \mathbf{x}_c ,\mathbf{y}_c, \mathbf{z}_c \right ] =\left [ \frac{l_{rcl-lcl} }{\left | l_{rcl-lcl} \right |} , \frac{\mathbf{z}_c\otimes \mathbf{x}_c}{\left | \mathbf{z}_c\otimes \mathbf{x}_c \right | } , \frac{l_{lownec-lcl} \otimes l_{lownec-rcl} }{\left | l_{lownec-lcl} \otimes l_{lownec-rcl} \right | }  \right ]  \\
         &\left [ \mathbf{x}_p ,\mathbf{y}_p, \mathbf{z}_p \right ] =\left [   \frac{\mathbf{y}_p \otimes l_{th-mspine} }{\left | \mathbf{y}_p \otimes l_{th-mspine} \right | } ,
         \frac{l_{th-lownec} }{\left | l_{th-lownec} \right |} ,\frac{\mathbf{x}_p\otimes \mathbf{y}_p}{\left | \mathbf{x}_p\otimes \mathbf{y}_p \right | }  \right ] 
        \end{aligned}~, \right.
\end{equation}

(8) thorax
\begin{equation}\label{eq: cpdir}
\left\{ \begin{aligned}
&\left [ \mathbf{x}_c ,\mathbf{y}_c, \mathbf{z}_c \right ] =\left [   \frac{\mathbf{y}_c \otimes l_{th-mspine} }{\left | \mathbf{y}_c \otimes l_{th-mspine} \right | } ,
         \frac{l_{th-lownec} }{\left | l_{th-lownec} \right |} ,\frac{\mathbf{x}_c\otimes \mathbf{y}_c}{\left | \mathbf{x}_c\otimes \mathbf{y}_c \right | }  \right ]  \\
         &\left [ \mathbf{x}_p ,\mathbf{y}_p, \mathbf{z}_p \right ] =\left [   \frac{\mathbf{y}_p \otimes l_{mspine-root} }{\left | \mathbf{y}_p \otimes l_{mspine-root} \right | } ,
         \frac{l_{mspine-th} }{\left | l_{mspine-th} \right |} ,\frac{\mathbf{x}_p\otimes \mathbf{y}_p}{\left | \mathbf{x}_p\otimes \mathbf{y}_p \right | }  \right ] 
        \end{aligned}~, \right.
\end{equation}

(9) mid spine
\begin{equation}\label{eq: cpdir}
\left\{ \begin{aligned}
&\left [ \mathbf{x}_c ,\mathbf{y}_c, \mathbf{z}_c \right ] =\left [   \frac{\mathbf{y}_c \otimes l_{mspine-root} }{\left | \mathbf{y}_c \otimes l_{mspine-root} \right | } ,
         \frac{l_{mspine-th} }{\left | l_{mspine-th} \right |} ,\frac{\mathbf{x}_c\otimes \mathbf{y}_c}{\left | \mathbf{x}_c\otimes \mathbf{y}_c \right | }  \right ]   \\
         &\left [ \mathbf{x}_p ,\mathbf{y}_p, \mathbf{z}_p \right ] =\left [  \frac{l_{rhip-lhip} }{\left | l_{rhip-lhip} \right |}, \frac{l_{root-rhip} \otimes l_{root-lhip} }{\left | l_{root-rhip} \otimes l_{root-lhip} \right | } 
          ,\frac{\mathbf{x}_p\otimes \mathbf{y}_p}{\left | \mathbf{x}_p\otimes \mathbf{y}_p \right | }  \right ] 
        \end{aligned}~, \right.
\end{equation}

(10) left hip 
\begin{equation}\label{eq: cpdir}
\left\{ \begin{aligned}
&\left [ \mathbf{x}_c ,\mathbf{y}_c, \mathbf{z}_c \right ] =\left [   \frac{\mathbf{y}_c\otimes \mathbf{z}_c}{\left | \mathbf{y}_c\otimes \mathbf{z}_c \right | } , 
         \frac{l_{lk-lhip} }{\left | l_{lk-lhip} \right |} , \frac{l_{lhip-rhip} \otimes -\mathbf{y}_c }{\left | l_{lhip-rhip} \otimes -\mathbf{y}_c \right | } \right ]   \\
         &\left [ \mathbf{x}_p ,\mathbf{y}_p, \mathbf{z}_p \right ] =\left [  \frac{l_{rhip-lhip} }{\left | l_{rhip-lhip} \right |}, \frac{l_{root-rhip} \otimes l_{root-lhip} }{\left | l_{root-rhip} \otimes l_{root-lhip} \right | } 
          ,\frac{\mathbf{x}_p\otimes \mathbf{y}_p}{\left | \mathbf{x}_p\otimes \mathbf{y}_p \right | }  \right ] 
        \end{aligned}~, \right.
\end{equation}

(11) left knee
\begin{equation}\label{eq: cpdir}
\left\{ \begin{aligned}
&\left [ \mathbf{x}_c ,\mathbf{y}_c, \mathbf{z}_c \right ] =\left [    \frac{-\mathbf{y}_c \otimes l_{lk-lhip} }{\left | -\mathbf{y}_c \otimes l_{lk-lhip} \right | },
         \frac{l_{lan-lk} }{\left | l_{lan-lk} \right |} ,  \frac{\mathbf{x}_c\otimes \mathbf{y}_c}{\left | \mathbf{x}_c\otimes \mathbf{y}_c \right | } \right ]   \\
         &\left [ \mathbf{x}_p ,\mathbf{y}_p, \mathbf{z}_p \right ] =\left [   \frac{\mathbf{y}_c\otimes \mathbf{z}_c}{\left | \mathbf{y}_c\otimes \mathbf{z}_c \right | } , 
         \frac{l_{lk-lhip} }{\left | l_{lk-lhip} \right |} , \frac{l_{lhip-rhip} \otimes -\mathbf{y}_c }{\left | l_{lhip-rhip} \otimes -\mathbf{y}_c \right | } \right ]
        \end{aligned}~, \right.
\end{equation}

(12) left ankle:
\begin{equation}\label{eq: cpdir}
\left\{ \begin{aligned}
&\left [ \mathbf{x}_c ,\mathbf{y}_c, \mathbf{z}_c \right ] =\left [    \frac{\mathbf{y}_c\otimes \mathbf{z}_c}{\left | \mathbf{y}_c\otimes \mathbf{z}_c \right |}, \frac{l_{lan-lk}}{\left | l_{lan-lk} \right | },
         \frac{l_{lan-lf} }{\left | l_{lan-lf} \right |}   \right ]   \\
         &\left [ \mathbf{x}_p ,\mathbf{y}_p, \mathbf{z}_p \right ] =\left [    \frac{-\mathbf{y}_p \otimes l_{lk-lhip} }{\left | -\mathbf{y}_p \otimes l_{lk-lhip} \right | },
         \frac{l_{lan-lk} }{\left | l_{lan-lk} \right |} ,  \frac{\mathbf{x}_p\otimes \mathbf{y}_p}{\left | \mathbf{x}_p\otimes \mathbf{y}_p \right | } \right ]
        \end{aligned}~, \right.
\end{equation}

(13) right hip
\begin{equation}\label{eq: cpdir}
\left\{ \begin{aligned}
&\left [ \mathbf{x}_c ,\mathbf{y}_c, \mathbf{z}_c \right ] =\left [   \frac{\mathbf{y}_c\otimes \mathbf{z}_c}{\left | \mathbf{y}_c\otimes \mathbf{z}_c \right | } , 
         \frac{l_{rk-rhip} }{\left | l_{rk-rhip} \right |} , \frac{-\mathbf{y}_c \otimes  l_{rhip-lhip}}{\left | -\mathbf{y}_c \otimes  l_{rhip-lhip} \right | } \right ]   \\
         &\left [ \mathbf{x}_p ,\mathbf{y}_p, \mathbf{z}_p \right ] =\left [  \frac{l_{rhip-lhip} }{\left | l_{rhip-lhip} \right |}, \frac{l_{root-rhip} \otimes l_{root-lhip} }{\left | l_{root-rhip} \otimes l_{root-lhip} \right | } 
          ,\frac{\mathbf{x}_p\otimes \mathbf{y}_p}{\left | \mathbf{x}_p\otimes \mathbf{y}_p \right | }  \right ] 
        \end{aligned}~, \right.
\end{equation}

(14) right knee
\begin{equation}\label{eq: cpdir}
\left\{ \begin{aligned}
&\left [ \mathbf{x}_c ,\mathbf{y}_c, \mathbf{z}_c \right ] =\left [    \frac{-\mathbf{y}_c \otimes l_{rk-rhip} }{\left | -\mathbf{y}_c \otimes l_{rk-rhip} \right | },
         \frac{l_{ran-rk} }{\left | l_{ran-rk} \right |} ,  \frac{\mathbf{x}_c\otimes \mathbf{y}_c}{\left | \mathbf{x}_c\otimes \mathbf{y}_c \right | } \right ]   \\
         &\left [ \mathbf{x}_p ,\mathbf{y}_p, \mathbf{z}_p \right ] =\left [   \frac{\mathbf{y}_p\otimes \mathbf{z}_p}{\left | \mathbf{y}_p\otimes \mathbf{z}_p \right | } , 
         \frac{l_{rk-rhip} }{\left | l_{rk-rhip} \right |} , \frac{-\mathbf{y}_p \otimes  l_{rhip-lhip}}{\left | -\mathbf{y}_p \otimes  l_{rhip-lhip} \right | } \right ]
        \end{aligned}~, \right.
\end{equation}

(15) right ankle
\begin{equation}\label{eq: cpdir}
\left\{ \begin{aligned}
&\left [ \mathbf{x}_c ,\mathbf{y}_c, \mathbf{z}_c \right ] =\left [    \frac{\mathbf{y}_c\otimes \mathbf{z}_c}{\left | \mathbf{y}_c\otimes \mathbf{z}_c \right |}, \frac{l_{ran-rk}}{\left | l_{ran-rk} \right | },
         \frac{l_{ran-rf} }{\left | l_{ran-rf} \right |}   \right ]   \\
         &\left [ \mathbf{x}_p ,\mathbf{y}_p, \mathbf{z}_p \right ] =\left [ \frac{-\mathbf{y}_p \otimes l_{rk-rhip} }{\left | -\mathbf{y}_p \otimes l_{rk-rhip} \right | },
         \frac{l_{ran-rk} }{\left | l_{ran-rk} \right |} ,  \frac{\mathbf{x}_p\otimes \mathbf{y}_p}{\left | \mathbf{x}_p\otimes \mathbf{y}_p \right | } \right ] 
        \end{aligned}~, \right.
\end{equation}
where $ls$, $le$, $lw$, $lcl$,$mspine$,$lownec$,$th$,$root$,$lhip$,$lk$,$lan$ and $lf$ represent left shoulder, left elbow, left wrist, left clavicle, mid spine, lower neck, thorax, root, left hip, left knee, left ankle and left foot.

\begin{figure*}[t]
\centering
\includegraphics[scale=0.98]{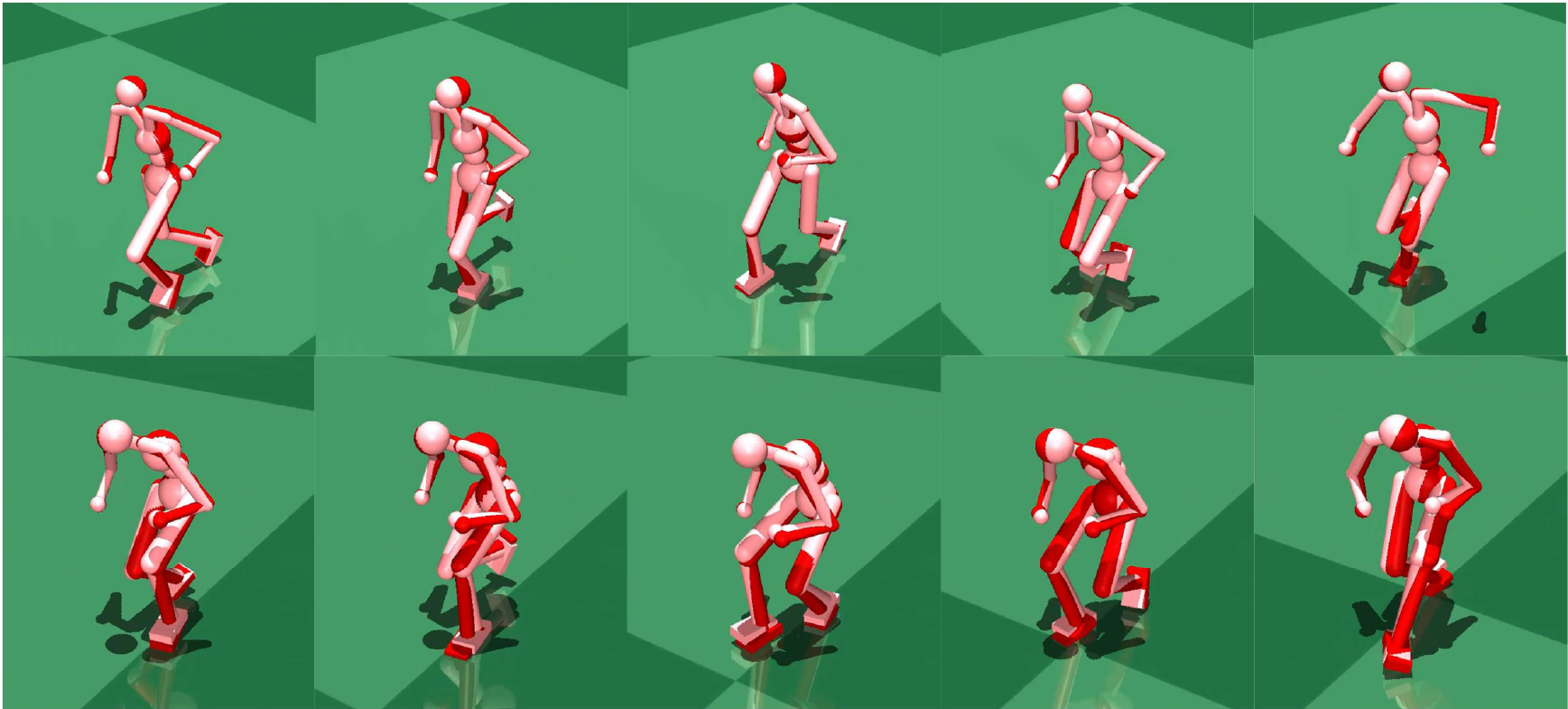}
\caption{Qualitative results of the reference motions ({\color{red}red humanoids}) and imitated motions ({\color{pink}pink humanoids}) by our curriculum residual force control policy $\pi_{CRP}$.}
\label{supp crf results}       
\end{figure*}

\section{Implementation Details}
\subsection{Network Architecture}
Our motion in-betweening network is the same as Harvey et al. \cite{harvey2020robust}, which is a popular motion in-betweening model based on recurrent neural network. Our curriculum residual force control policy $\pi_{CRP}$ is a multi-layer preceptron (MLP) network with 3 hidden layers whose size is (1024,512,256).

\subsection{Experiments Details}
We implement our whole system on PyTorch \cite{paszke2019pytorch}. The three stages in our Skeleton2Humanoid system are implemented in a sequential manner. The motion in-betweening network is trained following \cite{harvey2020robust} on the train set of the LaFAN1/Noobjects LaFAN1 dataset with various transition lengths (from 5
frames to 30 frames) between past keyframes and a future keyframe, then the test time adaptation stage individually optimizes the pretrained motion in-betweening network on different test sets whose transition lengths of skeleton motions are 5 frames, 15 frames and 30 frames. The skeleton to humanoid matching stage converts human skeleton motions synthesized from the test sets to humanoid motions. Our curriculum residual force control policy $\pi_{CRP}$ is trained on the converted humanoid motions from the test sets. We severally train a policy for the converted humanoid motions with different transition lengths (5 frames, 15 frames and 30 frames).

We utilize Mujoco as our physics simulator. During RL training, we apply the standard reinforcement learning algorithm PPO. The policy network is a typical 3-layer MLP network with size (1024, 512, 256). We train the curriculum residual force control policy for 1500 iterations. For each iteration, we keep collecting data by sampling RL episodes until the total number of time steps reaches 50000. For each RL episode, the RL agent explores to act and imitate a randomly sampled matched humanoid motion sequence and creates corresponding rewards. The RL episode is terminated when the end frame is reached or the humanoid’s root height is 0.5 below the root height of the reference pose.


\section{Qualitative Evaluation}

\subsection{Qualitative Results of Skeleton2Humanoid System}
We provide more visualization results of our Skeleton2humanoid System including ground truth skeleton motions, predicted skeleton motions, converted humanoid motions and imitated humanoid motions in attached videos. As shown in our videos, our approach can synthesize various physically-plausible humanoid motions.

\subsection{Qualitative Results of Our Curriculum Residual Force Control Policy}
In this section, we provide more qualitative results of our curriculum residual force policy $\pi_{CRP}$ (CRP) which are shown in Fig. \ref{supp crf results}. we can see that our curriculum residual force control policy can imitate many complex motions accurately.

\begin{table}
\begin{tabular}{lllllll}
&   & L2P &                         &   & L2Q &     \\ \hline
length              & 5 & 15  & \multicolumn{1}{l|}{30} & 5 & 15 & 30 \\ \hline
baseline\cite{harvey2020robust}   &  0.183 &  0.548  & \multicolumn{1}{l|}{1.167}   & 0.138  &  0.373  & 0.667 \\ \hline
iteration 1   & 0.171  & 0.538   & \multicolumn{1}{l|}{1.153}  & 0.135  &  0.370  & 0.663  \\ \hline
iteration 2   & 0.167  & 0.532   & \multicolumn{1}{l|}{1.145}   & 0.134  &  0.368  & 0.661 \\ \hline
iteration 3   & 0.164  & 0.529   & \multicolumn{1}{l|}{1.141}   & 0.134  &  0.367  & 0.661  \\ \hline
iteration 4   & 0.163  & 0.527   & \multicolumn{1}{l|}{1.140}  & 0.133  &  0.367  & 0.661  \\ \hline
iteration 5   & 0.162  & 0.526   & \multicolumn{1}{l|}{1.142}   & 0.133  &  0.367  & 0.663  \\ \hline
iteration 6   & 0.162  & 0.526   & \multicolumn{1}{l|}{1.146}   & 0.133  &  0.367  & 0.666  \\ \hline
iteration 7   & 0.162  & 0.527   & \multicolumn{1}{l|}{1.151}   &  0.133  &  0.367  & 0.669  \\ \hline
\end{tabular}
\caption{performance gains from test time adaptation and the number of iterations.}
\label{supp inference time}
\end{table}

\begin{table}
\begin{tabular}{llll}
&   & wall time &                     \\ \hline
length              & 5 & 15  & 30  \\ \hline 		
baseline\cite{harvey2020robust}   & 8.30ms  & 8.85ms  & 9.88ms   \\ \hline		
iteration 1   &0.91ms  &  1.75ms   & 3.08ms  \\ \hline
iteration 2   & 1.83ms  & 3.56ms  & 6.18ms  \\ \hline
iteration 3   & 2.75ms  & 5.39ms   & 9.30ms    \\ \hline
iteration 4   & 3.68ms  & 7.18ms   & 12.43ms  \\ \hline
iteration 5   & 4.61ms  & 8.97ms   & 15.54ms   \\ \hline
iteration 6   & 5.54ms  & 10.78ms   & 18.70ms   \\ \hline
iteration 7   & 6.46ms  & 12.57ms   & 21.83ms   \\ \hline
\end{tabular}
\caption{The relation between the inference/wall time (per sequence) and the number of iterations.}
\label{supp gain}
\end{table}

\subsection{inference time penalty of the TTA stage}
 We provide the trade-off results between the extra performance gains from TTA and the number of iterations as suggested by the reviewer in table. \ref{supp inference time} and table. \ref{supp gain}. The result is obtained on the LaFAN1 dataset (about 2245 sequences) with various transition lengths (5 frames, 15 frames, 30 frames) using a single 3090 GPU.

\end{document}